\documentclass[runningheads]{llncs}
\usepackage[T1]{fontenc}
\usepackage{graphicx}
\begin{document}
\title{Passing Heatmap Prediction Based on Transformer Model Using Tracking Data For Football Analytics}
%
%
\author{Yisheng Pei\inst{1}\orcidID{0000-0002-8245-5484} \and
Varuna De Silva\inst{1}\orcidID{0000-0001-7535-141X} \and
Mike Caine\inst{1,2}}
\authorrunning{Y. Pei et al.}
%
\institute{Institute for Digital Technologies, Loughborough University London, UK  \and
University of Warwick, UK}
\maketitle      
\begin{abstract}
Although the data-driven analysis of football players' performance has been developed for years, most research only focuses on the on-ball event including shots and passes, while the off-ball movement remains a little-explored area in this domain. Players’ contributions to the whole match are evaluated segmentally and unfairly, those who have more chances to score goals earn more credit than others, while the indirect and unnoticeable impact that comes from continuous movement has been ignored. This research presents a novel deep-learning network architecture which is capable to predict the potential end location of passes and how players’ movement before the pass affects the final outcome. Once analyzed more than 28,000 pass events, a robust prediction can be achieved with more than 0.7 Top-1 accuracies. And based on the prediction, a better understanding of the pitch control and pass option could be reached to measure players’ off-ball movement contribution to defensive performance. Moreover, this model could provide football analysts a better tool and metric to understand how players’ movement over time contributes to the game strategy and final victory.

\end{abstract}
\section{Introduction}
Data-based analysis of athletics performance has been developed for years, as the most complicated and popular sport around the world, the research on the performance of football players is being paid more and more attention~\cite{aalbers_2018_distinguishing,arbussangesa_2020_using,bransen_2019_choke}. Nowadays, with advances in technology and big data, we have several advanced metrics and methods to evaluate individual players’ contribution to the match outcome or goal-scoring opportunities, for instance, xG~\cite{decroos_2019_actions}, xT~\cite{singh_2018_introducing}, and many other relatives~\cite{decroos_2020_vaep,fernndez_2018_wide}. While most previous research only focuses on the on-ball event during the competition, including passes~\cite{arbussangesa_2020_using,goes_2021_a}, shot~\cite{goes_2020_the} and dribble, which ignores the fact that each player has only 3 minutes to control the ball on average as opposed to being free of the ball for 87 minutes. If players run randomly and spontaneously, the outcome will not likely be positive. Thus the decision-making capability during that 87 minutes without the ball determined whether they were really committed to the team's strategy and performance. This paper addresses the problem of estimating the movement decision capability when players are defending and propose a deep machine learning model that allows us to value and compare their decision with other potential choices. 

The measurement of decision-making ability is essentially a comparison value between the happened decision of players during defending and the potential choice based on all the given spatiotemporal information. As the ultimate goal of attacking, the purpose of all the on-ball events relates to creating goal-scoring chances. On the other side of the coin, the purpose of defending is to avoid the increase of opponent teams’ success probability. This paper takes the most frequent scene in a football match, passing, which covers over 80 per cent of the match events, as the research objective. To reach this level, we calculate the pass-end location possibilities and reflect the relationship with certain players’ movement together. 

We propose a self-attention mechanism model to learn the pass-end location possibilities depending on the defenders’ trajectories. In order to achieve the purpose of valuing off-ball movement, we regard the problem as a time series classification task while the end zone of each pass is the category that is being predicted. 

The main contributions of this work are the following:\\
- We propose a model for estimating the possible final outcome and the reality to evaluate football players’ movement decision-making capability while defending, which allows us to analyse players on a different and comprehensive level and include more match context. \\
- We develop a self-attention architecture to estimate how players’ movement affects the match process and shows that the deep learning model has much more potential for football performance analysis.\\
- We present a novel and practical metric which is able to help to identify outstanding defending players and improve their decision-making capability.

This paper is organised as follows: relevant literature in sports analysis and machine learning (ML) is presented in Section 2. The proposed xPass model is introduced in Section 3, along with all technical details. The results of experiments and potential application are shown in Section 4 and Section 5 respectively. Finally, in the last section, we summarise the paper and indicate the outlines of future work.

\begin{figure}
\centering
\includegraphics[width=0.5\textwidth]{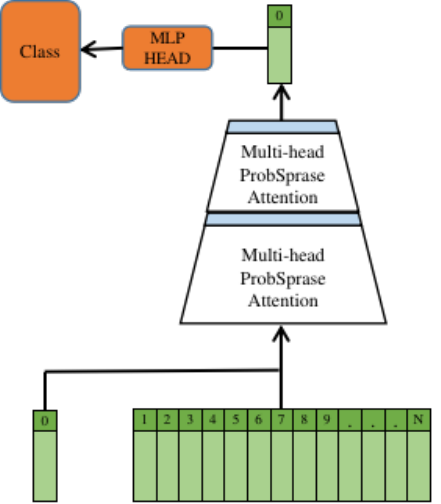}
\caption{The xPass Model Overview. The encoder receives sequential embedded tracking data (green series) with an extra learnable ‘classification token’~\cite{dosovitskiy_2022_an} for the classification task. Instead of the canonical self-attention block, we replace it with the ProbSparse self-attention from the Informer model~\cite{zhou_2021_informer} to avoid memory limitation.} \label{fig1}
\end{figure}

\section{Background}
Considering the practical and applied usage of xD metric, our work is inspired by several previous other approaches aimed at estimating~\cite{rudd_2011_a,teranishi_2022_evaluation,yam_2019_attacking} and predicting the game process based on the tracking~\cite{robberechts_2021_a,statsbomb_2021_introducing} and event data in the professional football domain. Technically, instead of the traditional convolutional neural networks (CNN) in computer vision tasks, we leverage the latest self-attention transformer model~\cite{vaswani_2017_attention} and treat the problem as a multi-class classification task.

\subsection{Football Performance Analysis}
The performance analysis of football players has been developed since the Moneyball theory~\cite{lewis_2004_moneyball}. Afterwards, Rudd~\cite{rudd_2011_a} firstly and creatively treated the football game as a Markov Chain, which quantifies and connects the actions of players with the goal-scoring probability. Since then various approaches have been built to achieve a similar goal, measure the contribution of players’ on-ball events to the final game result~\cite{mackay_2018_introducing}. Some of them treat the target straightly as the shot chances and quality, while others choose a more indirect way to measure the pitch control or heatmap on the pitch during the possession~\cite{martens_2021_space,shaw_2020_routine}. The technical methods applied in previous research differ, from the expert-guided development of algorithmic rules to a linear regression model with a set of handcrafted features~\cite{goes_2021_modelling}, from a CNN-based model~\cite{cheong_2021_prediction} to an assembled tree algorithm~\cite{fernndez_2019_decomposing}. Little previous researches put attention to the other side of the coin, the majority of game playing and players’ decision-making process, movement efficiency without the ball and how that choices discriminate smart players from normal ones.

\subsection{Self-Attention and Transformer}
The transformer model is initially designed to solve the sequential machine translation task between two languages~\cite{vaswani_2017_attention}. After the self-attention mechanism is approved to have strong feature representing and capturing capabilities, instead of recurrent neural network (RNN) series, it becomes the state-of-the-art in Natural Language Processing (NLP) domain. Based on its special characters, the transformer also shows a huge prospect in the image tasks including image classifications~\cite{dosovitskiy_2022_an}, object detection~\cite{liu_2021_swin} and semantic segmentation~\cite{lin_2021_a}. 

Besides the single task, as the development of this architect, the multi-task problems can be solved as well, which shares similar features with football performance analysis. They both have various paired and unpaired inputs with a number of types and the outputs have a certain flexibility due to the specific requests. Some early research has been attempted to extract features from the event dataset and football broadcasting video~\cite{simpson_2022_seq2event,zhou_2021_informer} and recognize the group activities in team sports~\cite{li_2022_learning,tamura_2022_hunting}. In our case, the transformer-based model is used to extract useful features from football tracking datasets and applied them to several practical tasks separately.

\section{Methodology}
Within this section, we propose a transformer-based time series deep learning model to estimate the possibility of the ball end location of any given pass based on prior information. To achieve the goal, the model is built on top of the tracking data collected from real football matches, containing the x and y location of all 22 players and the ball at 25 frames per second. 

In architectural design, we used a combined classification and time stamp embedding to deal with the pass data while a ProbSparse attention encoder extracts features for further tasks. Our proposed model aims to solve the pass prediction problem as a time-series classification problem. Please refer to Fig. 1 for an overview and check the following sections for detail. 

\subsection{Loss Function}
Defining a proper loss function that measures the gap between prediction and ground truth is necessary and crucial for a model. While the model output can be varied according to the tasks, not all the information is highly and directly associated with the target. For example, predicting the players who are top-3 most likely to receive the ball has been analysed before~\cite{arbussangesa_2020_using}, and the number of players has been utilized as the target feature normally. 

\begin{figure}
\centering
\includegraphics[width=0.7\textwidth]{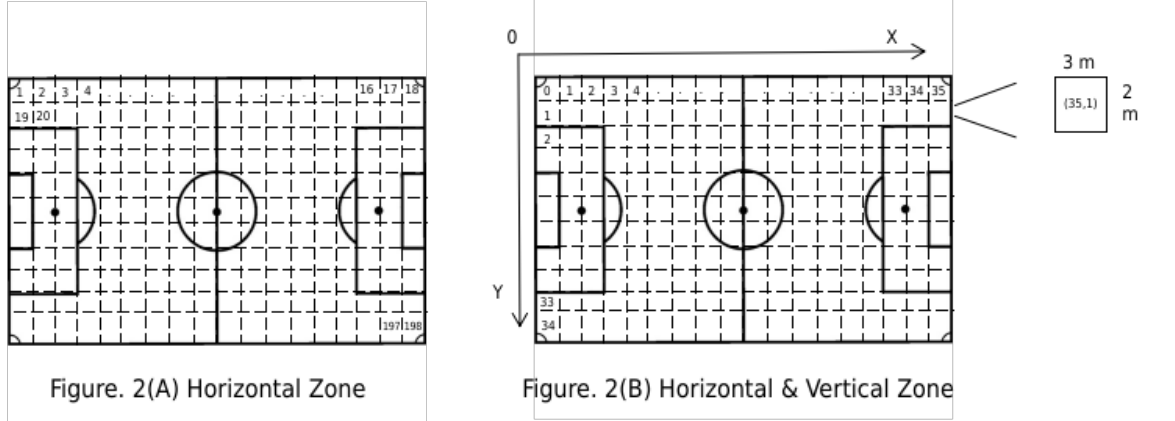}
\caption{The whole football pitch is divided into multi zones on both the x-axis and y-axis. Each zone contains the same area of the original pitch.} \label{fig2}
\end{figure}

Instead of predicting the potential ball receiver, which is a 9-classes classification problem, the ball passer and the goalkeeper are ignored. Due to the reality of football matches, most of the passes are aimed at the spare space rather than a staying player. Thus the understanding of pass location possibility on the whole pitch matters, because it is much closer to the game philosophy. While the standard pitch is 106 meters * 65 meters, how to split the pitch into a large number of zones affects the final result significantly. For example, if we have either a single horizontal or vertical zone on the pitch like Fig. 2(A), the total amount would be up to 1200 which is too many for football datasets and the number will lose geographical meaning as zone 1 is actually next to zone 19, but would be marked as the neighbour of zone 18, which is on the other end of the pitch. The problem could be solved by using a combined coordinate on the x-axis and y-axis separately. Eventually, the pitch has two sides, one has 35 zones on the x-axis and 34 zones on the y-axis, leading to each zone being 3 meters long and 2 meters wide and can be labelled as (35,1) in Fig. 2(B), another one has 105 zones on the x-axis and 68 zones on the y-axis, leading to each zone being 1 meter long and 1 meter wide.

We train the xPass model by minimizing the Cross-Entropy Loss (CEL) between the estimated and ground truth on both the x-axis and y-axis. The loss function ${L(\widehat{z},z)}$ is defined in Equation 1. 

\begin{equation}
L(\widehat{z},z) = CEL(\widehat{z_x},z_x) + CEL(\widehat{z_y},z_y)
\end{equation}

where ${\widehat{z_x}}$ and ${\widehat{z_y}}$ represents the prediction of the ball end location on the x-axis and y-axis separately while ${{z_x}}$ and ${{z_y}}$ means the ground truth.

\subsection{Model Architecture}
The xPass model comprises several stages according to different tasks. Inputs are the match tracking data from ${(t-2)}$ to ${t}$  seconds, with the output prediction of the pass event at time ${t}$. 

\subsubsection{Class token and Global embedding}
The input representation process includes classification token embedding and time-series global embedding. The classification token embedding is inspired by the Vision Transformer work~\cite{dosovitskiy_2022_an} and is alignable. While the time-series global embedding is required to capture the long-range sequential features and patterns for the task. 

Similar to BERT’s [class] token~\cite{devlin_2018_bert}, the first token of every sequence is always a special classification token which would be extracted after the attention blocks. A classification head is attached afterwards. In Fig. 1, the 0 green series which was added at the beginning and extracted after the attention blocks stands for the classification token. 

For the time-series global embedding, we have followed the method proposed by the Informer model~\cite{zhou_2021_informer}. Three separate parts are considered, a scalar projection, a local time stamp and a global time stamp. The scalar projection is aimed at aligning the dimension of inputs. While the local time stamp embedding is the same as the vanilla fixed position embedding, in order to solve the time sequential prediction problem, the global information (week, month and year) or other special time stamps like holidays and events would be considered. In the case of football, previous match results, recent fixtures, current goal differences etc. could be the factors.

After the input representation, the t-th sequence input ${X_t}$ would be shaped into a matrix ${X_{en}^t \in R^{(l_x+1) * d_{model}}}$.

\subsubsection{Efficient Self-Attention Mechanism}
Self-attention The canonical self-attention mechanism is initially proposed by the transformer paper~\cite{vaswani_2017_attention}. It adopts the Query-Key-Value (QKV) matrix calculation, given the packed matrix representations of queries ${Q \in R^{N*D_k}}$, keys ${K \in R^{M*D_k}}$ and values ${V \in R^{M*D_v}}$. The classical scaled dot-product attention score function is as follows: 
 
\begin{equation}
Attention(Q, K, V) = softmax(\frac{QK^T}{\sqrt{d_k}})V
\end{equation}
\\
where M and N represent the lengths of keys and queries; ${D_k}$  and ${D_v}$ denote the dimensions of keys and values; $A = Softmax(\frac{QK^T}{\sqrt{d_k}})$ is often called attention matrix; softmax is applied in a row-wise manner. The dot-products of queries and keys are divided by ${\sqrt{D_k}}$ to alleviate the gradient vanishing problem of the softmax function, which should be altered case by case.

\begin{figure}
    \centering
    \includegraphics[width=0.6\textwidth]{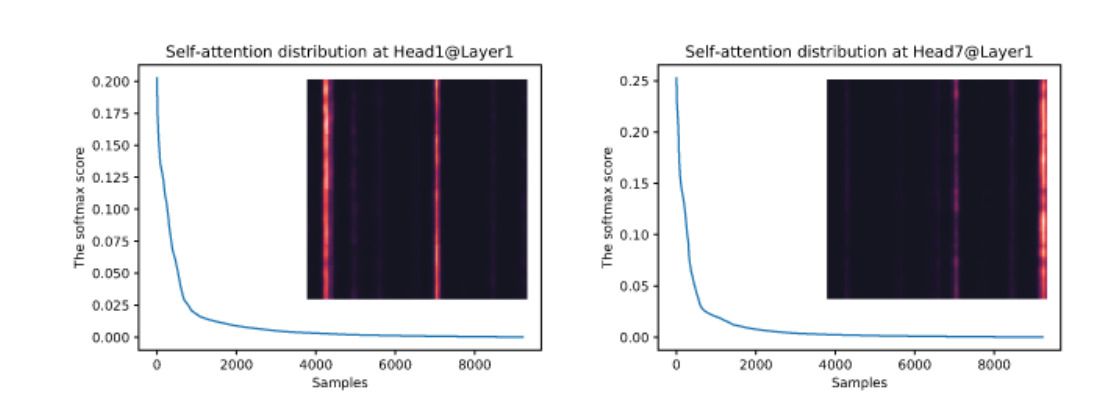}
    \caption{The Softmax scores in the self-attention from a 4-layer canonical Transformer trained on the weather datasets~\cite{zhou_2021_informer}.} \label{fig3}
\end{figure}

One of the significant advantages of the self-attention mechanism compared to other time-series algorithms like Recurrent-Neural-Network (RNN) variants~\cite{sherstinsky_2018_fundamentals} is that the calculation can be fully parallelized, whereas an RNN always processes the sentences sequentially. The inherent cost of this mechanism leads to a higher demand for memory usage on GPU. 

However, it has been proved that the distribution of self-attention probability has potential sparsity which means it forms a long tail distribution~\cite{tsai_2019_transformer}, and only a few dot-product pairs provide a contribution to the major attention score, while most of them are quick irrelevant (Fig. 3). 

In a more visual and intuitionistic case, as shown in Fig. 4, at this specific pass event moment, due to the domain knowledge of professional football, the position of players in the red boxes has a much more direct and important impact on the ball holder’s pass decision, while the players in the yellow boxes are supposed to affect less.

\begin{figure}
    \centering
    \includegraphics[width=0.7\textwidth]{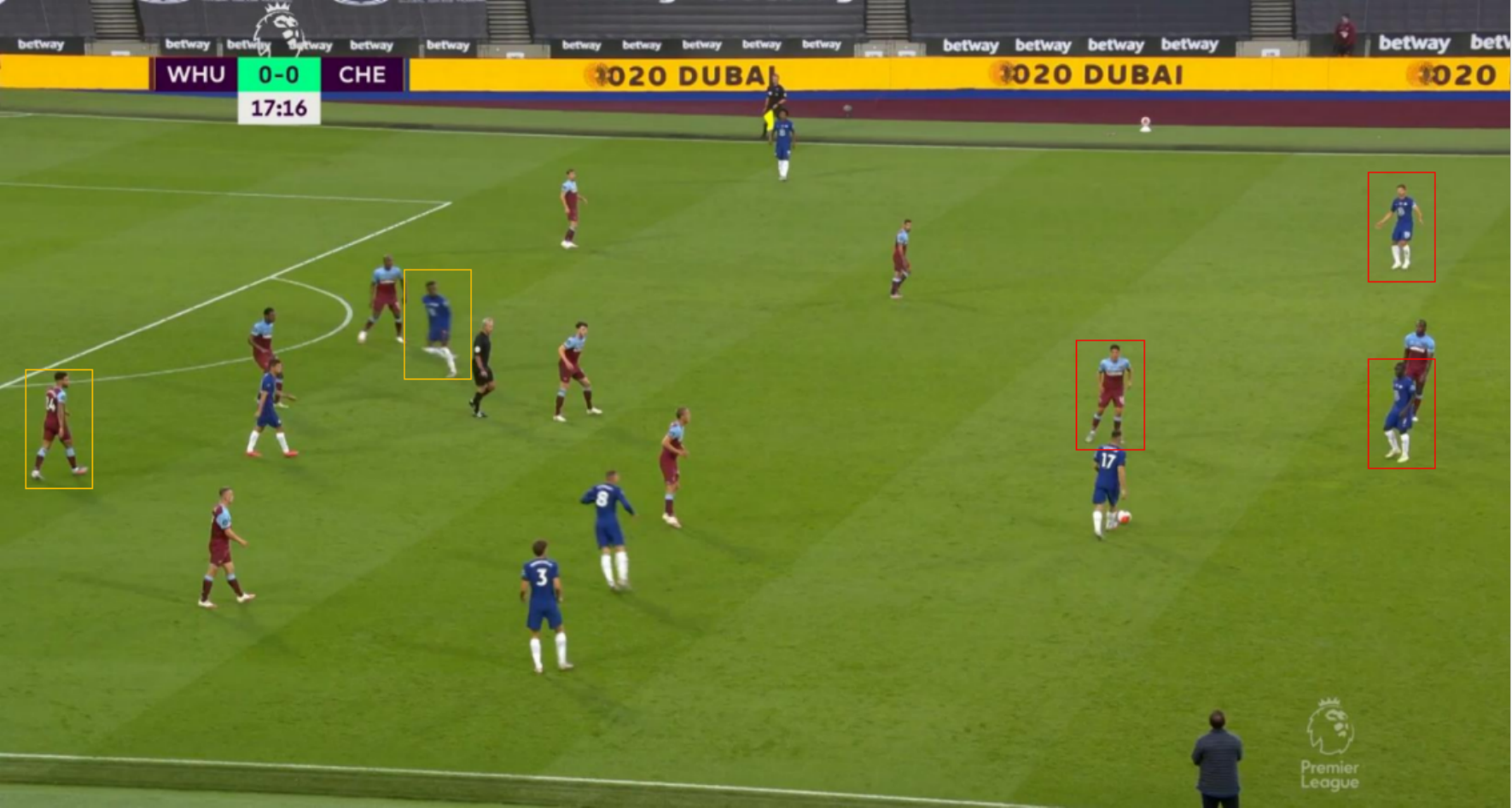}
    \caption{The moment before a pass event happens on the pitch.} \label{fig4}
\end{figure}

In order to distinguish the real contributed dot-product pairs from other lazy pairs, we follow the ProbSparse Self-attention mechanism from the Informer model~\cite{zhou_2021_informer}. The measurement consists of two parts. The first part is the query sparsity measurement, which compares the corresponding query’s attention probability distribution with the uniform distribution through Kullbakc-Leibler (KL) divergence as the dominant dot-product pairs are supposed to differ from the uniform one while other lazy pairs tend to be similar. Second, based on the above query sparsity measurement, each key in the ProbSparse Self-attention is not attend to all queries but only to a specific amount of dominant pairs, which can be adjusted by a sampling factor. Eventually, the time complexity and space complexity of the ProbSparse Self-attention drops from ${O(L^2)}$ to ${O(L ln L)}$ significantly, where L stands for the input length of queries and keys.

The ProbSparse Self-attention equation is shown as:

\begin{equation}
Attention(Q, K, V) = softmax(\frac{\overline{Q}K^T}{\sqrt{d_k}})V
\end{equation}

where ${\overline{Q}}$ is a sparse matrix of the same size as the original ${Q}$ matrix, which only contains the top dominant queries rather than all of the inputs. 

\begin{figure}
    \centering
    \includegraphics[width=0.7\textwidth]{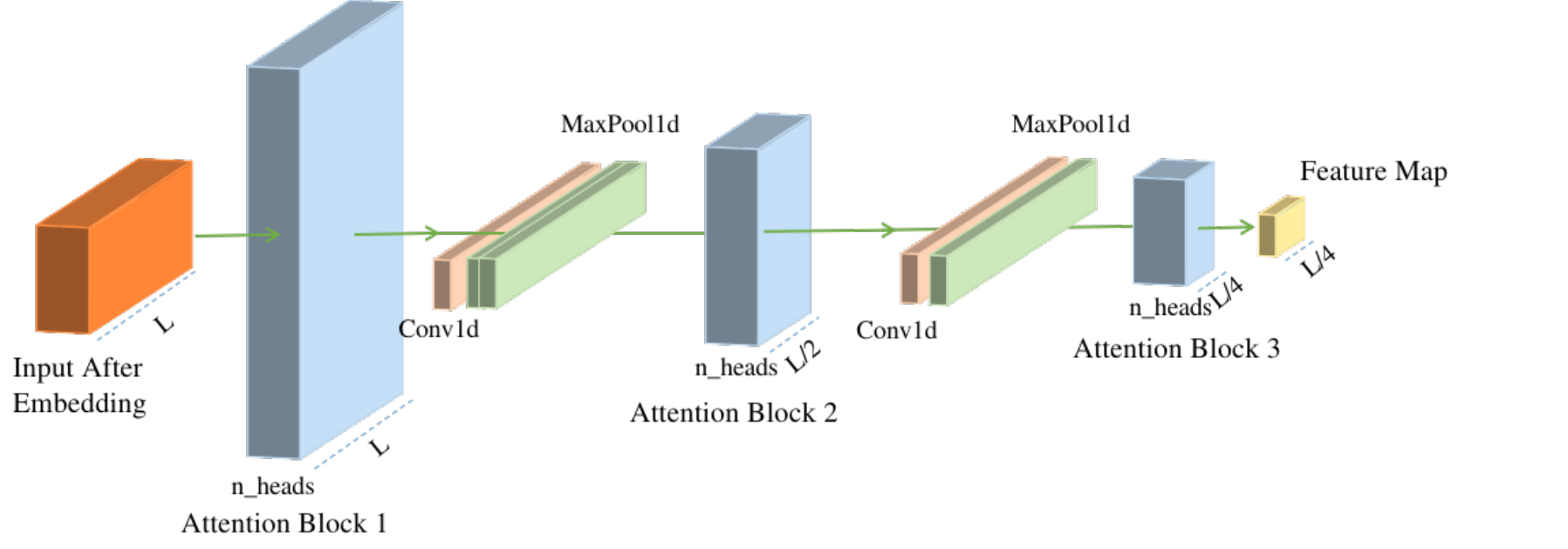}
    \caption{The moment before a pass on the pitch.} \label{fig5}
\end{figure}

\subsubsection{Encoder Stack for Sequential Inputs}

The encoder is designed to extract robust and representative feature maps and is composed of a stack of N = 2 identical encoders. We give a sketch of the encoder in Fig. (5) for clarity. 

Each encoder has N = 3 ProbSparse Attention Block and a distilling operation connects each of these attention blocks. Due to the initially designed purpose, only a few dominated queries contribute the most are selected, while the redundancy should be trimmed. The distilling operation consists of a 1-D Convolutional filter layer with a kernel width = 3, a Max-Pooling layer with stride 2 and an Exponential Linear Unit (ELU) activation function. The distilling operation forward from ${i}$-th distilling layer to ${(i+1)}$-th layer is followed:

\begin{equation}
    X_{i+1}^t = MaxPool(ELU(Conv1d([X_i^t]_{A_B})))
\end{equation}
                
where ${[X_i^t]_A{}_B}$ is the t-sequence input ${X^t}$ after the attention block.  
Since then the output dimension of the encoder stack is aligned, which is hinged on the number of encoder stacks and attention blocks. All the stacks’ outputs would be concatenated at the end as the final output feature map of the encoder.

\section{Experiment}
In this section, we describe the datasets we used to train and validate our model with the testing performance of the proposed architecture for the pass-end location possibility estimation. 

\subsection{Datasets}
This research is based on the tracking and event data generated from 30 Premier League matches of the 2019/2020 and 2020/2021 seasons, provided by a professional football club. The tracking data for each game contains the (x,y,z) location for each individual player and the ball at 25Hz. The event data provides the time, event type, football team id, player involved, event start location, event end location, the outcome of the event and the body part used. In total, 16 different events are collected including pass, dribble, throw in and shot. From the datasets, we extract 28,872 passing events, which are split into a training, validation and test set with a 70:10:20 distribution. The validation set is used for model selection during a grid-search process while the result is based on the test dataset. Similar tracking and event dataset are available at {https://github.com/metrica-sports/sample-data}.

\subsection{Experimental Details}
We used the Adam optimizer with 1 = 0.9 and  2 = 0.999 for the proposed methods and started the learning rate from 1e-4 with a proper early stopping setting. The dimension of the transformer model is 512 and the numbers of the multi-head are 8. Two encoder stacks consist of the main encoder, while each of them has three ProbSparse self-attention blocks with two distilling connections. We trained the model on a NVIDIA RTX A6000 GPU with a batch size of 32.

\begin{figure}
    \centering
    \includegraphics[width=0.7\textwidth]{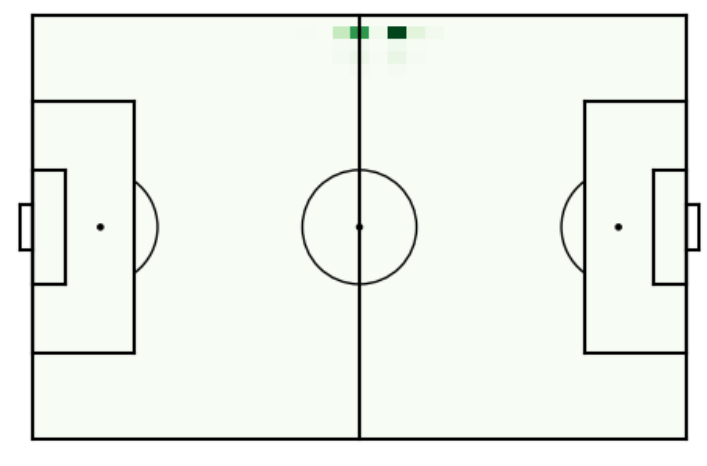}
    \caption{The moment before a pass on the pitch.} \label{fig6}
\end{figure}
Fig. 6 shows the prediction heatmap of the passing end location, the darker colour zone represents a higher possibility to become the destination of the pass choice while the red point stands for the ground truth end location of that decision.

We also compare our model with three previous related works~\cite{arbussangesa_2020_using,Honda,soccermap}. The SoccerMap model is based on the heatmap of the tracking data with a combined computer vision and handcrafted features method to estimate the passing probability on the whole pitch, which consists of thousands of cells that approximately stand for ${1 m^2}$ on the football pitch. We also cite two additional model results from their work, both of which are built on a set of handcrafted features generated from the tracking data. One is called \textit{Logistic Net} which is a network with a single sigmoid unit, and another is \textit{Dense2 Net} which is a neural network with two dense layers with a ReLu activation layer and a sigmoid output unit.

The \textit{Pass Feasibility Model} is designed to predict the potential pass receiver based on the players’ orientation, location and opponents’ spatial configuration. They treat the problem as a 9-classes classification problem, excluding the ball holder and the goalkeeper. While another work \textit{Pass Receiver Model} by~\cite{Honda}, applied a computer vision method to the same task, comparing the difference between whether the visual information from the broadcasting video matters or not. 

\subsection{Results and Analysis}

\textit{Table 1} presents the results of the proposed xPass model for the pass-end location prediction problem and three previous similar target works mentioned above.

\begin{table}[]
\centering
\caption{The result of the pass end location prediction with three relative models .}\label{tab1}
\begin{tabular}{|c|c|c|c|c|}
\hline
Model                                                                           & CEL Loss       & \begin{tabular}[c]{@{}c@{}}Top-1 \\ Accuracy(\%)\end{tabular} & \begin{tabular}[c]{@{}c@{}}Top-3 \\ Accuracy(\%)\end{tabular} & \begin{tabular}[c]{@{}c@{}}Top-5 \\ Accuracy(\%)\end{tabular} \\ \hline
\begin{tabular}[c]{@{}c@{}}the xPass Model \\ (2 * 3)\end{tabular}              & \textbf{0.249} & N/A                                                            & N/A                                                            & N/A                                                            \\ \hline
\begin{tabular}[c]{@{}c@{}}the xPass Model \\ (1 * 1)\end{tabular}              & \textbf{0.277} & N/A                                                            & N/A                                                            & N/A                                                            \\ \hline
Logistic Net~\cite{soccermap}                                                                    & 0.384          & N/A                                                            & N/A                                                            & N/A                                                            \\ \hline
Dense2 Net~\cite{soccermap}                                                                      & 0.349          & N/A                                                            & N/A                                                            & N/A                                                            \\ \hline
SoccerMap~\cite{soccermap}                                                                       & \textbf{0.217} & N/A                                                            & N/A                                                            & N/A                                                            \\ \hline
\begin{tabular}[c]{@{}c@{}}Pass Feasibility \\ Model~\cite{arbussangesa_2020_using}\end{tabular}               & 0.624          & 37.6                                                         & 71.0                                                         & N/A                                                            \\ \hline
\begin{tabular}[c]{@{}c@{}}Pass Receiver Model\\  (without vision)~\cite{Honda}\end{tabular} & 0.510          & 49.0                                                         & 84.9                                                         & 95.0                                                         \\ \hline
\begin{tabular}[c]{@{}c@{}}Pass Receiver Model\\  (with vision)~\cite{Honda}\end{tabular}    & 0.375          & 62.5                                                         & 92.3                                                         & 97.5                                                         \\ \hline
\end{tabular}
\end{table}

We convert the Top-1 Accuracy from~\cite{Honda} to the CEL loss as the comparison metric. However, their task is only a 9-classes classification problem while ours have up to 7,140 zones, which is on another level. Even though, our model overperforms those works, whether they are built on handcrafted features or a deep neural network with tracking and vision datasets. On the other hand, although the purpose of the SoccerMap model differs from ours, we both have a similar definition of the pitch zone and extend the pass event from a player-to-player relation but to a player-to-space event. We applied two totally different methods but match an even result, their work is based on the images generated from the tracking datasets and they focus on a certain passing moment while the time-series information before the moment is ignored. We believe that should not be missed whether from either a machine learning perspective or a football performance analysis perspective. That leads to the final result for us, without any handcrafted features but still achieving good performance on the task. 

\section{Conclusion}
The novel xPass model applies the latest ProbSparse self-attention sequential machine learning techniques to predict the pass outcome and how it is affected by the other player’s decision. The passing possibility heatmap is built on the model, which reflects players’ off-ball movement contribution to defending during opponents’ possession. 
In terms of practical application, the model has the capability of measuring the off-ball movement impact on the opponents' passing decisions and pitch control ability. The model based on the spatiotemporal tracking data can be transferred and applied to other similar sports, for many other tasks, it could also be the benchmark processing on the tracking and event data. We suggest more expert knowledge should be involved for a more detailed scenario analysis while more 3D body information will bring the model much closer and more realistic to the game situation.

%
%
%

\end{document}